\documentclass[11pt]{article}
\usepackage{coling2020}
\usepackage{times}
\usepackage{url}
\usepackage{latexsym}

\usepackage[T1]{fontenc}
\usepackage[utf8]{inputenc}
\usepackage{microtype}
\usepackage{graphicx}
\usepackage{subcaption}
\usepackage{array}
\usepackage{booktabs}
\usepackage{multirow}
\usepackage{rotating}
\usepackage{makecell}
\usepackage{longtable}
\usepackage{colortbl}
\usepackage{multicol}
\usepackage{float}
\usepackage{placeins}

\graphicspath{{figures/}{../figures/}{../../figures}}

\newcolumntype{H}{>{\setbox0=\hbox\bgroup}c<{\egroup}@{}}
\newcolumntype{P}[1]{>{\centering\arraybackslash}p{#1}}
\newcommand\Tstrut{\rule{0pt}{9pt}}
\newcommand\Bstrut{\rule[-4pt]{0pt}{0pt}}

\def\mypar#1{\vspace{2mm}{\noindent\bfseries #1.}\hspace{1mm}}

\newcommand{\ul}[1]{\bf\underline{#1}}

\usepackage{transparent}

\usepackage{mathtools}
\usepackage{amssymb} % widetilde
\usepackage{mathrsfs} % mathscr
\usepackage{nameref}  % reference sections by name

\def\fig#1{Figure~\ref{fig:#1}}

\def\tab#1{Table~\ref{tab:#1}}

\def\sect#1{\textsection\ref{sec:#1}}

\usepackage{xspace}  % for onedot
\usepackage[inline]{enumitem}
\setlist[description]{leftmargin=\parindent,labelindent=\parindent, itemsep=4pt, parsep=4pt}
\newlist{observations}{enumerate*}{10}
\setlist[observations]{%
  label=(\bf\arabic*),
  itemjoin=\ ,
}

\definecolor{p1blue}{RGB}{0, 90, 200}     % blue
\definecolor{p1red}{RGB}{170, 10, 60}      % raspberry
\definecolor{p1green}{RGB}{10, 155, 75}    % green
\definecolor{p1yellow}{RGB}{234, 214, 68}  % yellow
\definecolor{p1orange}{RGB}{255, 130, 95}  % vermillion
\definecolor{p1purple}{RGB}{130, 20, 160}  % purple
\definecolor{p1azure}{RGB}{0, 160, 250}    % azure

\definecolor{p2blue}{RGB}{2, 29, 181}      % rgb(2, 29, 181)
\definecolor{p2red}{RGB}{137, 0, 0}        % rgb(137, 0, 0)
\definecolor{p2green}{RGB}{4, 79, 19}      % rgb(4, 79, 19)

\definecolor{padgray}{rgb}{.7,.7,.7}         % rgb(50, 50, 50) 
\definecolor{dimgray}{rgb}{.35,.35,.35}      % rgb(35, 35, 35)
\definecolor{darkgray}{rgb}{.20,.20,.20}     % rgb(20, 20, 20)

\definecolor{p3blue}{RGB}{27, 23, 255}       % rgb(27, 23, 255)
\definecolor{p3red}{RGB}{187,0,33}      
\definecolor{p3purple}{RGB}{200, 151, 185}   
\definecolor{p4blue}{RGB}{15, 75, 255}       % rgb(15, 75, 255)

\usepackage{tikz, pgfplots}
\usepackage{pgffor,pgfmath}

\pgfplotsset{
    compat=1.15,
    grid style={darkgray},
    minor grid style={dimgray!20, line width=0.1pt},
    major grid style={dimgray!20},
    axis line style = { darkgray }, 
    every axis plot/.append style={line width=1.5pt, mark options=solid, mark size=4pt},
    legend style={draw = darkgray, rounded corners=0pt, fill = white, font=\Large},
    tick style ={color = dimgray!30 },
    tick label style={font=\normalsize},
    label style={font=\normalsize},
}

\usetikzlibrary{decorations, shapes, arrows, arrows.meta, positioning, decorations.pathreplacing}
\usetikzlibrary{pgfplots.groupplots, patterns}
\usetikzlibrary{pgfplots.statistics}
\usepgfplotslibrary{external, fillbetween}

\pgfplotscreateplotcyclelist{systemspalette}{%
    smooth, darkgray, mark=o, mark size=3pt, line width=1.1pt, mark options=solid\\%PA-offline
    smooth, darkgray, mark=square, mark size=3pt,line width=1.1pt, mark options=solid\\%PA-online
    smooth, p4blue, mark=o, mark size=3pt, line width=1.1pt, mark options=solid\\%TF-offline
    smooth, p4blue, mark=square, mark size=3pt, line width=1.1pt, mark options=solid\\%TF-online
    }

\tikzset{
    txt/.style={ % style for observed variables
      text=darkgray,
      font=\large,
      align=center,
      inner sep=2pt,
      outer sep=2pt
    },
    read/.style={ 
      -{Straight Barb[angle=60:1pt 2]}, line width=1.1pt,
        p4blue
    },
    write/.style={
      -{Straight Barb[angle=60:1pt 2]}, line width=1.1pt,
        black
    },
    cote/.style={
      {Straight Barb[angle=60:1pt 2]}-{Straight Barb[angle=60:1pt 2]}, line width=.8pt,
        darkgray
    },
    lightconn/.style={  % connection
      -{Straight Barb[angle=60:1pt 2]}, line width=.5pt,
        darkgray
    },
    conn/.style={
      -{Straight Barb[angle=60:1pt 2]}, line width=1.1pt,
        darkgray
    }}

\newcommand\ulc[3]{{% #1 color #2 text #3 label -2pt the rule's yshift and 1.5pt its thickness
  \setbox0=\hbox{\textcolor{black}{\transparent{1}#2}{\transparent{1}\scriptsize(#3)}}
  \color{#1}\transparent{0.7}\ooalign{\copy0\cr\rule[\dimexpr-2pt\relax]{\wd0}{1.5pt}}}}

\newcommand{\waitk}{\mbox{\emph{wait-$k$}}}
\newcommand{\ld}{lagging difficulty}
\newcommand{\Ld}{Lagging difficulty}
\newcommand{\engde}{En$\sto$De}
\newcommand{\deen}{De$\sto$En}

\newcommand{\cond}{\,|\,}
\newcommand{\z}{\boldsymbol z}
\newcommand{\y}{\boldsymbol y}
\newcommand{\hyp}{{\tilde\y}}
\newcommand{\conf}{{p(\hyp|\x)}}

\newcommand{\x}{\boldsymbol x}

\newcommand{\unk}{{<}unk{>}}

\newcommand{\lx}{{|\boldsymbol x|}}
\newcommand{\ly}{{|\boldsymbol y|}}

\newcommand{\lhyp}{{|\tilde\y|}}

\newcommand{\ktr}{{k_\text{train}}}
\newcommand{\kev}{{k_\text{eval}}}

\newcommand{\sto}[1][5pt]{\mathrel{%
   \hbox{\rule[\dimexpr\fontdimen22\textfont2-.2pt\relax]{#1}{.4pt}}%
   \mkern-4mu\hbox{\usefont{U}{lasy}{m}{n}\symbol{41}}}}
\newcommand{\toto}{\leftrightarrow}

\makeatletter
\DeclareRobustCommand\onedot{\futurelet\@let@token\@onedot}
\def\@onedot{\ifx\@let@token.\else.\null\fi\xspace}

\def\eg{\emph{e.g}\onedot} 
\def\ie{\emph{i.e}\onedot}

 \def\vs{\emph{vs}\onedot} 

\def\wrt{w.r.t\onedot} 

\makeatother

\def\tabvspace1{\vspace{-8mm}}
\def\tabvspace2{\vspace{-2mm}}
\def\figvspace1{\vspace{-25mm}}
\def\figvspace2{\vspace{-5mm}}

\colingfinalcopy % Uncomment this line for the final submission

\title{Online Versus Offline NMT Quality: An In-depth Analysis on English–German and German–English}

\author{Maha Elbayad$^{1,2}$\quad
    Michael Ustaszewski$^{3}$\quad
    Emmanuelle Esperan\c{c}a-Rodier$^{1}$\\
    \bf
   Francis Brunet Manquat$^{1}$\quad
   Jakob Verbeek$^{4}$\quad Laurent Besacier$^{1}$\\
    $^1$LIG - Universit{\'e} Grenoble Alpes, France \quad $^2$Inria - Grenoble, France \\
    $^3$ University of Innsbruck, Department of Translation Studies \quad
 $^4$ Facebook AI Research  
}

\date{}

\begin{document}
\maketitle
\begin{abstract}
We conduct in this work an evaluation study comparing offline and online neural machine translation architectures. 
Two sequence-to-sequence models: convolutional Pervasive Attention~\cite{Elbayad18conll} and attention-based Transformer~\cite{Vaswani17nips} are considered. 
We investigate, for both architectures, the impact of online decoding constraints on the translation quality through a carefully designed human evaluation on English-German and German-English language pairs,  the latter being particularly sensitive to latency constraints. 
The evaluation results allow us to identify the strengths and shortcomings of each model when we shift to the online setup.
\end{abstract}

\section{Introduction}\label{sec:intro}
Sequence-to-Sequence models are state-of-the-art in a variety of sequence transduction tasks including machine translation (MT).
The most widespread models are composed of an encoder that reads the entire source sequence, while a decoder (often equipped with an attention mechanism) iteratively produces the next target token given the full source and the decoded prefix. 
Aside from the conventional \emph{offline} use case, recent works adapt sequence-to-sequence models for \emph{online} (also referred to as \emph{simultaneous}) decoding with low-latency constraints~\cite{Gu17eacl,Dalvi18naacl,Ma19acl,Arivazhagan19acl}.
Online decoding is desirable for applications such as real-time speech-to-speech interpretation.
In such  scenarios, the decoding process starts before the entire input sequence is available, and online prediction generally comes at the cost of reduced translation quality.

\begin{figure}
    \centering
    \includegraphics[height=3.4cm]{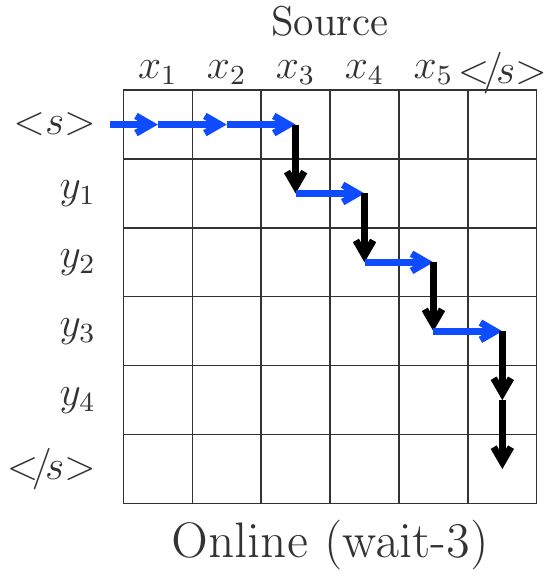}\hskip12pt
    \includegraphics[height=3.4cm]{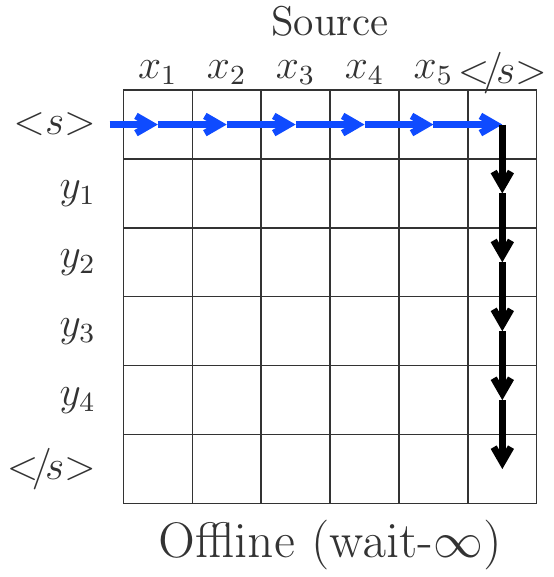}
\vspace{-3mm}
    \caption{Wait-$k$ decoding as a sequence of reads (horizontal) and writes (vertical) over a source-target grid. After first reading $k$ tokens, the decoder alternates between reads and writes. In Wait-$\infty$, or Wait-until-End (WUE), the entire source is read first.
}\label{fig:waitk-path}
\vspace{-4mm}
\end{figure}

In this work we focus on online neural machine translation (NMT) with deterministic \waitk~decoding policies~\cite{Dalvi18naacl,Ma19acl}.
With such a policy, we first read $k$ tokens from the source then alternate between producing a target token and reading another source token (see \fig{waitk-path}).
We consider two sequence-to-sequence models, a position-based convolutional model and a content-based model with self-attention. We specifically use the recent convolutional Pervasive Attention~\cite{Elbayad18conll} and Transformer~\cite{Vaswani17nips}. We investigate, for both architectures, the impact of online decoding constraints on the translation quality through a carefully designed human evaluation on English$\sto$German and German$\sto$English language pairs.

Our contributions are twofold: 
(1) our work, to the best of our knowledge, is the first human evaluation of online \vs offline NMT systems. 
(2) We compare Transformer and Pervasive Attention 
architectures highlighting the advantages and shortcomings of each when we shift to the online setup.
The rest of this paper is organized as follows: we present in \sect{related} related work pertaining to online MT and error analysis of NMT systems. 
We describe our experimental setup for human evaluation and error analysis in \sect{setup}. 
We follow with the evaluation results in \sect{results}
and summarize our findings in \sect{conclusion}.

\section{Related work}
\label{sec:related}

\subsection{Online NMT}

After pioneering works on online statistical MT (SMT) \cite{Fugen07MT,Yarmohammadi13naacl,He15emnlp,Grissom14emnlp,Oda15acl}, one of the early works with attention-based online translation is \newcite{Cho16arxiv} using manually designed criteria that dictate whether the model should make a read/write operation.
\newcite{Dalvi18naacl} proposed a deterministic decoding algorithm that starts with $k$ read operations then alternates between blocks of $l$ write/read operations.
This simple approach outperforms the information based criteria of \newcite{Cho16arxiv}, and allows complete control of the translation delay.
\newcite{Ma19acl} trained Transformer models \cite{Vaswani17nips} with a \waitk~decoding policy that first reads $k$ source tokens then alternate single read-writes.
For \emph{dynamic} online decoding, \newcite{Luo17icassp} and \newcite{Gu17eacl} rely on Reinforcement Learning to optimize a read/write policy.
To combine the end-to-end training of \waitk~models with the flexibility of dynamic online decoding, \newcite{Zheng19acl} and \newcite{Zheng19emnlp} use Imitation Learning.
Recent work on dynamic online translation use monotonic alignments \cite{Raffel17icml} with either a limited or infinite lookback \cite{Chiu18iclr,Arivazhagan19acl,Ma20iclr}. Another adjacent research direction enables revision during online translation to alleviate decoding constraints \cite{Niehues16interspeech,Zheng20acl,Arivazhagan20iwslt}.
In this work, we focus on \waitk~and greedy decoding strategies, but unlike 
other \waitk~models \cite{Ma19acl,Zheng19acl,Zheng19emnlp} we opt for uni-directional encoders which are efficient to train in an online setup \cite{Elbayad20interspeech}. 

\subsection{Error analysis for NMT}

With the advances in NMT~\cite{Bahdanau15iclr,Vaswani17nips},
the quality of translations has improved substantially leading to claims of human parity in high-resource settings ~\cite{Wu16arxiv,Hassan18arxiv}.
With such improvements, it becomes more and more difficult for automatic evaluation metrics such as BLEU~\cite{Papineni02acl} to detect subtle differences. 
Manual error annotation is a more instructive
quality assessment to gain insights into the performance of MT systems, especially in direct comparisons. Human evaluation might lead to conclusions at odd with automatic metrics, as was the case in last year's WMT English-German evaluation~\cite{Barrault19wmt}.

\mypar{Comparison to SMT and rule-based MT}
\newcite{Bentivogli16emnlp} studied post-editing of English-German TED talks 
and found that NMT makes considerably less word order errors than SMT.
They also observed that the performance of NMT degrades faster than SMT with increasing sentence length. 
\newcite{Toral17eacl} reached similar conclusions on news stories in 9 language directions.
\newcite{Isabelle17emnlp} tested NMT systems with \emph{challenging} linguistic material and highlighted the efficiency of NMT systems at handling subject-verb agreement and syntactic and lexico-syntactic divergences and the struggle of NMT with idiomatic phrases.
\newcite{CastilhoEtAl2017}, \newcite{Castilho2017ACQ} and \newcite{VanBrussel18lrec} observed that NMT outperforms SMT in terms of fluency, but at the same time it is more prone to accuracy errors. 
\newcite{Klubicka2018} made similar observations in an evaluation of English-Croatian, concluding that compared to SMT and rule-based MT, NMT tends to sacrifice completeness of translations in order to increase fluency.

\mypar{Error typologies for MT}
Various error typologies with different levels of granularity have been proposed to evaluate MT systems~\cite{Flanagan94amta,Vilar06lrec,Stymne12lrec,Lommel14mqm}.  In their evaluation of SMT outputs, \newcite{Vilar06lrec} defined five error categories: missing words, word order, incorrect words, unknown words and punctuation errors.
\newcite{Bentivogli16emnlp} followed a simpler classification with three types of errors: morphological, lexical, and word order.
Their choice was motivated by the difficulty to disambiguate sub-categories of lexical errors
\cite{Popovic11cl}.
The evaluation in \newcite{Klubicka2018} is based on the Multidimensional Quality Metrics (MQM) framework~\cite{Lommel14mqm}. 
In their study, they found that mistranslation is the most frequent accuracy error in NMT translations.
\newcite{VanBrussel18lrec} observed that mistranslation and omission errors are particularly challenging for NMT users, because contrary to SMT and rule-based MT, these errors are often not indicated by flawed fluency, which makes them more difficult to identify and post-edit. 

\mypar{Error analysis for online MT} 
\newcite{Hamon09eacl} evaluated a spoken language translation system (ASR+MT) in comparison to a human interpreter, where each segment is judged in terms of adequacy and fluency.
\newcite{Mieno15interspeech}, in search for a unique evaluation metric, examined the usefulness of delay and accuracy in predicting the human judgment of a simultaneous speech translation system.
To our knowledge, our work is the first to propose a fine-grained human evaluation of online NMT systems. It focuses on English$\sto$German and German$\sto$English language pairs, the latter being particularly sensitive to latency constraints. This is in part due to German sentence-final structures (\eg verbs in subordinate clauses) that require long-distance reordering in translation into syntactically divergent languages.

\section{Experimental setup}\label{sec:setup}
In this work we train Transformer~\cite{Vaswani17nips} and Pervasive Attention~\cite{Elbayad18conll} models for the tasks of online and offline translation. Following \newcite{Elbayad20interspeech}, 
we use unidrectional encoders and train the online MT models with $\ktr=7$, proven to yield better translations across the latency spectrum.
We train our models on IWSLT'14 \deen~(German$\sto$English) and \engde~(English$\sto$German) datasets~\cite{Cettolo14iwslt}.
Sentences longer than 175 words and pairs with length-ratio exceeding 1.5 are removed. 
The training set consists of 160K pairs with 7283 held out for development and the test set has 6750 pairs from TED dev2010+tst2010-2013.
All data is tokenized using the standard scripts from the Moses toolkit~\cite{Koehn07acl}. Unlike existing work experimenting with this dataset, we did not lowercase the bitexts so that we can correctly assess typography errors in German.
We segment sequences using byte pair encoding~\cite{Sennrich16acl}, BPE for short, on the bi-texts resulting in a shared vocabulary of 32K types.
We train Pervasive Attention (PA) with 14 layers and 7-wide filters and Transformer (TF) \emph{small} for offline and online translation.
We evaluated our \waitk~models with $\kev=3$ achieving a low latency of $\textrm{AL}\in [2.5, 3.5]$~(see \tab{automatic}).
TF models have 2M more parameters compared to PA (19M to 17M), they are however faster to train (PA is 8 times slower). In test time, the two models decode in comparable speeds.
For a fair comparison, both online and offline models are decoded greedily.
We will refer these four models with PA-offline, PA-online, TF-offline and TF-online.

\subsection{Analysis factors}\label{sec:factors}
In this section, we  describe the factors we use  to analyze the results of automatic and human evaluations.

\mypar{Source length}
Similar to other evaluation studies of NMT systems~\cite{Bentivogli16emnlp,Toral17eacl,Koehn17wnmt}, we look into the length of the source sequence and its effect on the quality of translation.

\mypar{\Ld~(LD)}
In the particular context of online translation, source-target alignments are an indicator of how \emph{easy} it is to translate an input.

To measure the \ld~of a pair $(\x, \y)$, we first estimate source-target alignments with \texttt{fast-align}~\cite{Dyer13naacl} and then infer a reference decoding path.
he reference decoding path, denoted with $\z^\text{align}$, is non-decreasing and guarantees that at a given decoding position $t$, $z_t$ is larger than or equal to all the source positions aligned with $t$.
The \ld~is finally measured as the Average Lagging (AL)~\cite{Ma19acl} of the parsed $\z^\text{align}$
as follows:
\begin{align}
    \textrm{LD}(\x,\y) =  \textstyle\frac{1}{\tau}\sum_{t=1}^\tau z_t^\text{align} -\frac{\lx}{\ly} (t-1),\quad
    \tau = \arg\min_t\{t \cond z_t = \lx\}.
\end{align}
AL measures the lag in tokens behind the ideal simultaneous policy wait-0, and so, LD measures the lag of a \emph{realistic} simultaneous translation that has the aligned context available when decoding.
The higher LD, the more challenging it is to constrain the latency of the translation.

Relying on alignments to assess a pair's difficulty is, however, not ideal; \newcite{Sridhar13naacl} experienced poor accuracies in streaming speech translation when segmenting the input based on alignments and \newcite{Grissom14emnlp} argued that the translator can accurately predict future words from a partial context and consequently beat the alignment-induced latency.

\mypar{Relative positions}
We look into the correlation between the relative positions, source-side and target-side, with the translation's quality. An annotated token $\tilde y_t$ of the system's hypothesis $\hyp$ has a target-side relative position $t/\ly$. Similarly, an annotated source token $x_j$ has a relative position $j/\lx$. 
We argue that with \waitk~decoding policies, the position of the token might be a contributing factor to the adequacy/fluency of the translation.

\begin{figure}
    \centering
    \includegraphics[width=.8\linewidth]{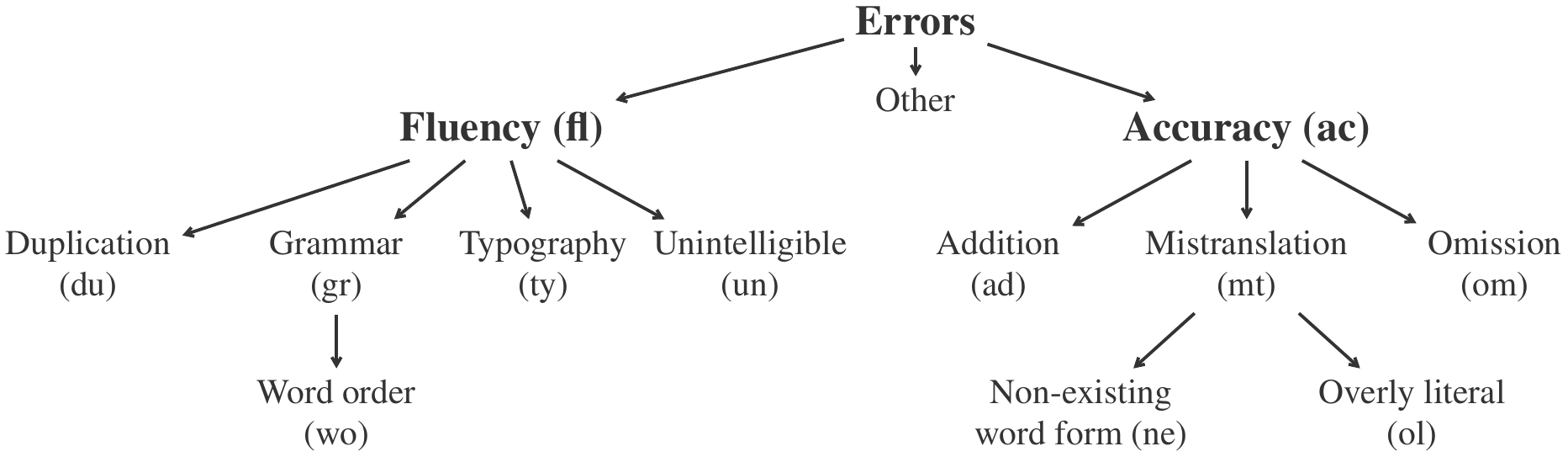}
\vspace{-3mm}
    \caption{MQM-based error typology used for our manual annotation. 
    }\label{fig:errortypology}
\vspace{-4mm}
\end{figure}

\subsection{Human evaluation}\label{sec:human}
In addition to the use of automatic evaluation metrics, we conduct an in-depth manual analysis to compare the quality of the output produced by the four systems. From the full test sets, we sample 200 segments 
in each translation direction.
First, we only keep segments whose source sentence lengths fall between the first and third length quartiles.
We then remove segments that contain the \unk~token (out-of-vocabulary) and bin the segments by \ld~(see \sect{factors}). We sample from the binned segments to cover all ranges of difficulty and manually remove misaligned segments. Subsequently, the sampled segments were manually error-annotated by a total of four human annotators.

\mypar{Error typology}\label{sec:typology}
For error annotation, a subset of the MQM error typology~\cite{Lommel14mqm} was used. 
A pilot annotation based on 50 translation segments not included into the present test data was carried out to select MQM error types relevant to this study. 
In this way, the typology was kept to a manageable size to avoid annotators' cognitive overload.\footnote{\url{http://www.qt21.eu/downloads/MQM-usage-guidelines.pdf}, \textsection~2.1} 
The resulting error typology comprises 13 error types, grouped into three major branches: \emph{accuracy}, \emph{fluency} and \emph{other}, as shown in \fig{errortypology}. 
The error type \emph{non-existing word form} was added to the typology 
to capture target words \emph{invented} with BPE tokenization that do not exist in the target language, such as translating \emph{Pfadfinder} (German for \emph{scout}) into \emph{Badfinder}.
The hierarchical nature of the error typology enables annotation and quality analyses on various levels of granularity; annotators were requested to give preference to more specific error types (\ie located deeper in the hierarchy) whenever possible.

\mypar{Annotation interface}
In order to annotate the segments, we use ACCOL\'E, an online collaborative platform for error annotation~\cite{esperancarodier:hal-02363208}.
ACCOL\'E offers a range of services that allow simplified management of corpora and error typologies with the possibility to specify the error typology and search for a particular error type in the annotations. 
Annotators are tasked to label translation errors by locating the appropriate spans in the target and source segments. 

\mypar{Selection and training of annotators}
Per language pair, two annotators with native proficiency in the respective target and near-native to native proficiency in the source language were recruited for the manual annotation task. 
Following the MQM guidelines
and recommendations from NMT evaluation studies~\cite{Laubli20jair}, the annotators are professional translators. 
Two of the annotators (one per language pair) also teach in a translation degree programme at university and have thus considerable experience in linguistic translation analysis. 
To familiarize the annotators with the annotation scheme and the use of ACCOL\'E, they were provided with training materials consisting of a written annotation manual, a description of the error typology and a decision tree to guide the selection of appropriate error types. 
In addition, annotators practiced the annotation procedure on a calibration set of 30 segments representative of the full test data but not included in the 200 segments to be annotated. 
Subsequently, annotators were given individual feedback and corrective guidance on their annotations.
The training materials are made publicly available for reuse.\footnote{
\url{https://github.com/elbayadm/OnlineMT-Evaluation}}
Annotators were remunerated 220 Euros each.

\section{Evaluation results}\label{sec:results}
\subsection{Automatic evaluation}\label{sec:automatic}
\begin{table}
    \centering
    \setlength{\tabcolsep}{2pt}
    \small
    \begin{tabular}{lrrr|rrr|rrr|rrr}
\toprule
& \multicolumn{6}{c|}{\deen} 
& \multicolumn{6}{c}{\engde}\\
\hline
& \multicolumn{3}{c|}{PA} & \multicolumn{3}{c|}{TF} 
& \multicolumn{3}{c|}{PA} & \multicolumn{3}{c}{TF}\\
& Offline & Online & $\Delta\%$ 
& Offline & Online & $\Delta\%$ 
& Offline & Online & $\Delta\%$ 
& Offline & Online & $\Delta\%$ \Bstrut\\
\hline
% Statistical significance with sample_size=3000, repetitions=100
% De-En
% PA-offline: [30.55, 31.24, 31.86]
% TF-offline: [30.47, 31.16, 31.89]
% Offline Proba PA > TF: 0.66  Proba TF > PA: 0.34
% PA-online: [25.75, 26.42, 27.17]
% TF-online: [25.93, 26.56, 27.26]
% Online  Proba PA > TF: 0.33  Proba TF > PA: 0.67
% En-De
% PA-offline: [25.30, 26.04, 26.76]
% TF-offline: [25.72, 26.59, 27.28]
% Offline Proba PA > TF: 0.02   Proba TF > PA: 0.98
% PA-online: [22.38, 23.02, 23.69]
% TF-online: [22.35, 22.98, 23.52]
% Online  Proba PA > TF: 0.55   Proba TF > PA: 0.45
$\uparrow$BLEU &\bf31.24 & 26.44 & -15  
               & 31.13 & \bf26.57 & -15 
               & 26.03 & \bf23.04 & -11 
               & \bf\ul{26.60} & 22.98 & -14 
               \Tstrut\\
% De-En: meteor-1.5-wo-en-no_norm-0.85_0.2_0.6_0.75-ex_st_sy_pa-1.0_0.6_0.8_0.6
% PA-offline: [28.5316, 28.9364, 29.3227]
% TF-offline: [28.8359, 29.2440, 29.6714]
% Offline Proba PA > TF: 0.01   Proba TF > PA: 0.99
% PA-online: [25.5947, 25.9687, 26.3316] 
% TF-online: [25.2839, 25.6545, 26.0861]
% Online  Proba PA > TF: 1.   Proba TF > PA: 0.0
% En-De: meteor-1.5-wo-de-no_norm-0.95_1.0_0.55_0.55-ex_st_pa-1.0_0.8_0.2
% PA-offline: [38.4254, 38.9777, 39.6658]
% TF-offline: [38.7908, 39.5047, 40.1318]
% Offline Proba PA > TF: .01  Proba TF > PA: 0.99
% PA-online: 35.2078, 35.7836, 36.3588]
% TF-online: a[34.9221, 35.4979, 36.1626]
% Online  Proba PA > TF: 0.96   Proba TF > PA: 0.04

$\uparrow$METEOR & 28.95 & \bf\ul{25.97} & -10
                 & \bf\ul{29.25} & 25.65  & -12
                 & 38.81 & \bf\ul{35.72}  & -8 
                 & \bf\ul{39.37} & 35.35 & -10 \\
% De-En
% PA-offline: [0.5530, 0.5640, 0.5773]
% TF-offline: [0.5445, 0.5553, 0.5665]
% Offline Proba PA > TF: 1.0   Proba TF > PA: 0.
% PA-online: [0.6089, 0.6218, 0.6355]
% TF-online: [0.6253, 0.6368, 0.6469]
% Online  Proba PA > TF: 0.   Proba TF > PA: 1.0
% En-De
% PA-offline: [0.61, 0.63, 0.64]
% TF-offline: [0.61, 0.62, 0.63]
% Offline Proba PA > TF: .86  Proba TF > PA: 0.14
% PA-online: [0.67, 0.68, 0.69]
% TF-online: [0.68, 0.69, 0.70]
% Online  Proba PA > TF: 0.   Proba TF > PA: 1.
$\downarrow$TER &  0.564 & \bf\ul{0.621} & +10 
                &  \bf\ul{0.555} & 0.637 & +16 
                & 0.626 & \bf\ul{0.676} & +8 
                & \bf0.622 & 0.692 & +11 \\
% DeEn
% PA-offline : [0.61, 0.62, 0.63]
% TF-offline : [0.62, 0.62, 0.63]
% Proba PA > TF: 0.07 Proba TF > PA: 0.93
% PA-online: [0.58, 0.58, 0.59]
% TF-online: [0.58, 0.59, 0.59]
% Online Proba PA > TF: 0.01 Proba TF > PA: 0.99
% EnDe
% PA-offline: [0.57, 0.58, 0.59]
% TF-offline: [0.58, 0.58, 0.59]
% Offline Proba PA > TF: 0.04 Proba TF > PA: 0.96
% PA-online: [0.55, 0.55, 0.56]
% TF-online: [0.55, 0.55, 0.56]
%Online Proba PA > TF: 0.52 Proba TF > PA: 0.48
$\uparrow$ROUGE-L& 62.89 & 59.30 & -6 
                 & \bf63.15 & \bf\ul{59.51} & -6 
                 & 57.99 & 55.41 & -4 
                 & \bf\ul{58.27} & \bf55.46 & -5 \\
% De-En
% PA-offline: [0.9363, 0.9374, 0.9384]
% TF-offline: [0.9374, 0.9384, 0.9397]
% Offline Proba PA > TF: 0.  Proba TF > PA: 1.0
% PA-online: [0.9263, 0.9277, 0.9288]
% TF-online: [0.9289, 0.9300, 0.9311]
% Online  Proba PA > TF: 0.  Proba TF > PA: 1.
% En-De
% PA-offline: [0.8538, 0.8564, 0.8589]
% TF-offline: [0.8549, 0.8575, 0.8595]
% Offline Proba PA > TF: 0.04   Proba TF > PA: 0.96
% PA-online: [0.8466, 0.8487, 0.8511]
% TF-online: [0.8457, 0.8481, 0.8505]
% Online  Proba PA > TF: 0.8   Proba TF > PA: 0.2
$\uparrow$BERTScore & 0.937 & 0.928 & -1 
                    & \bf\ul{0.939} & \bf\ul{0.930} & -1 
                    & 0.856 & 0.848 & -1 
                    & \bf\ul{0.858} & 0.848 & -1 \\
\midrule
AL & 21.10 & 2.59 & -88 
   & 21.10 & 3.16 & -85 
   & 20.71 & 3.33 & -84 
   & 20.71 & 3.49 & -83 \\
%ratio ($\lhyp/\ly$)  & 0.99 & 0.99 & +0 % TODO put back with comments
                     %& 0.96 & 1.06 & +10 
                     %& 1.01 & 1.06 & +5 
                     %& 1.00 & 1.08 & +8 \\
\bottomrule
\end{tabular}

%❯ bert-score -r deen_all/references.txt -c deen_all/pa_offline.txt --lang en
%roberta-large_L17_no-idf_version=0.3.0(hug_trans=2.4.1) P: 0.937303 R: 0.937776 F1: 0.937475

%❯ bert-score -r deen_all/references.txt -c deen_all/pa_online.txt --lang en
%roberta-large_L17_no-idf_version=0.3.0(hug_trans=2.4.1) P: 0.926759 R: 0.928838 F1: 0.927719

%❯ bert-score -r deen_all/references.txt -c deen_all/tf_offline.txt --lang en
%roberta-large_L17_no-idf_version=0.3.0(hug_trans=2.4.1) P: 0.939321 R: 0.937871 F1: 0.938537

%❯ bert-score -r deen_all/references.txt -c deen_all/tf_online.txt --lang en
%roberta-large_L17_no-idf_version=0.3.0(hug_trans=2.4.1) P: 0.928410 R: 0.931867 F1: 0.930073

%❯ bert-score -r ende_all/references.txt -c ende_all/pa_offline.txt --lang de
%bert-base-multilingual-cased_L9_no-idf_version=0.3.0(hug_trans=2.4.1) P: 0.858783 R: 0.854541 F1: 0.856434

%❯ bert-score -r ende_all/references.txt -c ende_all/pa_online.txt --lang de
%bert-base-multilingual-cased_L9_no-idf_version=0.3.0(hug_trans=2.4.1) P: 0.848178 R: 0.849135 F1: 0.848437

%❯ bert-score -r ende_all/references.txt -c ende_all/tf_offline.txt --lang de
%bert-base-multilingual-cased_L9_no-idf_version=0.3.0(hug_trans=2.4.1) P: 0.860221 R: 0.855343 F1: 0.857543

%❯ bert-score -r ende_all/references.txt -c ende_all/tf_online.txt --lang de
%bert-base-multilingual-cased_L9_no-idf_version=0.3.0(hug_trans=2.4.1) P: 0.846435 R: 0.850003 F1: 0.847992

\vspace{-2mm}
    \caption{Automatic evaluation of the full test sets. The better scoring system is in bold,  underline indicates that the system is better than its competitor with at least 95\% statistical significance.}\label{tab:automatic}
\vspace{-4mm}
\end{table}

\begin{figure}
    \centering
    \begin{subfigure}[b]{.45\linewidth}
        \centering
        \includegraphics[scale=1]{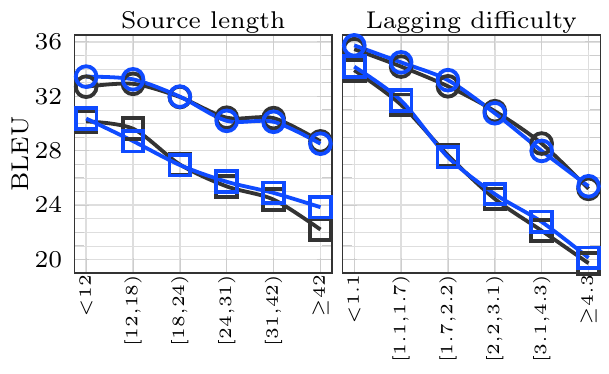}%
        \vspace{-5pt}
        \caption{\deen}
    \end{subfigure}\hfill
    \begin{subfigure}[b]{.55\linewidth}
        \centering
        \includegraphics[scale=1]{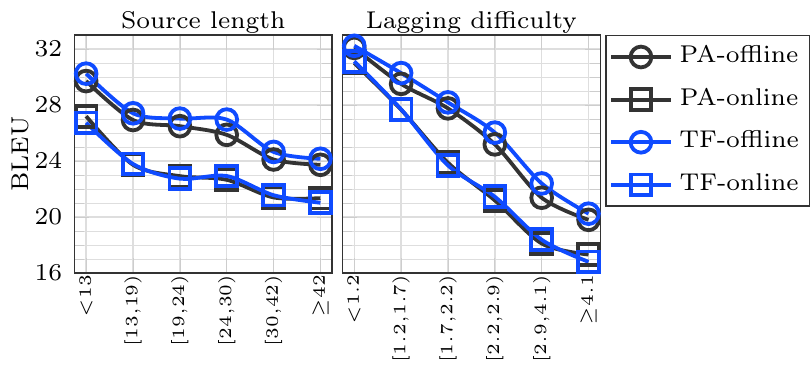}%
        \vspace{-5pt}
        \caption{\engde}
    \end{subfigure}
\vspace{-8mm}
    \caption{Bucketed BLEU scores by source length and by \ld~of the full test set.}\label{fig:bleu-buckets}
\vspace{-4mm}
\end{figure}

For each translation direction, \engde~and \deen, we assess the quality of our systems by measuring BLEU~\cite{Papineni02acl}, 
METEOR~\cite{Lavie07meteor},
TER~\cite{Snover06amta},
ROUGE-L~\cite{Lin14rouge}
and BERTScore~\cite{Zhang20iclr}. 
We use the default weights and parameters for METEOR\footnote{English: meteor-1.5-wo-en-no\_norm-0.85\_0.2\_0.6\_0.75-ex\_st\_sy\_pa-1.0\_0.6\_0.8\_0.6,\\\phantom{abcd}German: meteor-1.5-wo-de-no\_norm-0.95\_1.0\_0.55\_0.55-ex\_st\_pa-1.0\_0.8\_0.2} and
we report the F1 measure combining BERTScore precision and recall.\footnote{English: roberta-large\_L17\_no-idf\_version=0.3.0(hug\_trans=2.4.1),\\\phantom{abcd}German:  bert-base-multilingual-cased\_L9\_no-idf\_version=0.3.0(hug\_trans=2.4.1).} We test for statistical significance with paired bootstrap resampling~\cite{Koehn04emnlp} using a sample size of 3000 segments.
We report the automatic scores evaluated on IWSLT'14 De$\toto$En test set in \tab{automatic} and \fig{bleu-buckets}. For bucketed BLEU scores, we bin the test data based on the \ld~of the pair or the source length and measure corpus-level BLEU in each bin. 

We observe that
\begin{observations}
\item In offline translation, TF and PA have a comparable performance on \deen~with a slight advantage to TF on all metrics except from BLEU. In the \engde~direction, TF widens the gap with PA significantly.
When binning \engde~by \ld, PA is outperformed by TF in all ranges of difficulty except from the first \emph{easy} bin.
\item As to be expected, online decoding leads to a degradation of the translation quality. 
The degradation is higher for \deen~(5 BLEU points) than for \engde~(3 BLEU points), arguably because German uses not only verb-initial but also verb-final constructions depending on clause type, thus posing more latency-related challenges for online translation. 
\item When switching to online translation, the degradation of PA is narrower on average than the degradation of TF allowing for PA to close the gap with TF in both directions.
\item Although the translation quality of the systems in both directions decreases \wrt the length of the source segment, the length is a weaker feature for \engde~compared to \deen.
    \Ld~proves to be a better feature, not only in online translation, but also in offline translation with a steeper decline in BLEU scores as we increase the difficulty.
\end{observations} 

\begin{table}
\centering
\small
\setlength{\tabcolsep}{2pt}
\begin{tabular}{rlrrr|rrr|rrr|rrr}
\toprule
& & \multicolumn{6}{c|}{De$\sto$En}
& \multicolumn{6}{c}{En$\sto$De}\\
\hline
& & \multicolumn{3}{c|}{PA} & \multicolumn{3}{c|}{TF}
& \multicolumn{3}{c|}{PA} & \multicolumn{3}{c}{TF} \\

\hline
& System & Offline & Online & $\Delta\%$ & Offline & Online & $\Delta\%$
& Offline & Online & $\Delta\%$ & Offline & Online & $\Delta\%$\\
\hline
(ac)&Accuracy &   2 &   1 & -50 &   2 &   3 &  +50 
              &   0 &   0 &  +0 &   0 &   1 & - \\

(ad)&Addition &  \bf76 & \bf143 &  +88 &  95 & 160 &  +68 
              &  30 &  66 & +120 &  35 &  97 & +177 \\

(mt)&Mistranslation & \bf433 & 587 &  +36 & 457 & \bf572 &  +25 
                    & 245 & 260 &   +6 & \bf202 & 260 &  +29 \\

(ne)&Non-existing WF &  26 &  17 & -35 &  \bf14 &  \bf16 &  +14 
                     & \bf39 & 58 & +49 & 43 & \bf54 & +26 \\

(om)&Omission &  \bf67 & \bf113 &  +69 &  96 & 127 &  +32 
              &  \bf99 &  \bf74 & -25 & 126 & 114 & -10 \\

(ol)&Overly litteral & 78 & 95 & +22 & \bf52 & \bf81 & +56 
                     & 150 & 179 &  +19 & \bf113 & \bf125 &  +11 \\

\hline
(ac+)&Total accuracy & \bf682 & \bf956 &  +40 & 716 & 959 &  +34 
                     & 563 & \bf637 &  +13 & \bf519 & 651 &  +25
                     \Tstrut\Bstrut \\
\hline
(fl)&Fluency & 17 & 20 & +18 & \bf14 & 20 & +43 
             & 26 &  \bf21 & -19 &  \bf20 &  24 &  +20 \\

(du)&Duplication &  11 &  32 & +191 &  22 & 144 & +555 
&   5 &  15 & +200 &  13 &  71 & +446 \\

(gr)&Grammar & 57 & 65 & +14 & \bf36 & \bf34 & -6 
             & 198 & 260 &  +31 & \bf142 & \bf222 &  +56 \\

(ty)&Typography & 41 & \bf42 & +2 & \bf33 & 59 & +79 
                & 52 & 92 & +77 & \bf49 & \bf78 & +59 \\

(un)&Unintelligible  &  2 &  3 & +50 &  2 &  2 &  +0 
                     &  \bf4 &  \bf8 & +100 & 11 & 11 &  +0 \\

(wo)&Word order & \bf65 & 105 &  +62 &  66 &  \bf78 &  +18 
                & 46 & 85 & +85 & \bf37 & \bf74 & +100 \\

\hline
(fl+)&Total fluency & 193 & \bf267 &  +38 & \bf173 & 337 &  +95 
                    & 331 & 481 &  +45 & \bf272 & 480 &  +76 
                    \Tstrut\Bstrut\\
\hline
(ac+fl)&Total &  \bf875 & \bf1223 & +40 & 889 & 1296 &   +46 
              &  894 & \bf1118 &   +25  & \bf791 & 1131 &   +43 \Tstrut\Bstrut\\
\bottomrule
\end{tabular}

\vspace{-2mm}
\caption{Total number of errors (sum of two annotations) per error type for each system.
The system (PA or TF) with less errors is put in bold.
}\label{tab:raw}
\vspace{-4mm}
\end{table}

\subsection{Human evaluation}\label{sec:annotation}
To analyse the annotation data, we rely on the sum count of errors reported by the two annotators. For token-level analysis, we parse the span of each reported error and consider the union of the two annotations to label each output token.
To assess the reliability of the error annotation, we measure inter-annotator agreement (IAA) with Cohen’s $\kappa$~\cite{Cohen1960} at the token level measuring whether the two annotators agree on the exact error type assigned to each token. We observe an agreement of 0.33 for \deen~and 0.40 for \engde~which is compatible with other MQM-based evaluation studies~\cite{Lommel14eamt,Specia17mt}.

For each error type in our typology, we report in \tab{raw} the count of its occurrences as labeled by two annotators. The frequencies are arranged by task (\deen~and \engde), by system (PA and TF) and by decoding setup (offline and online). We observe that
\begin{observations}
\item In alignment with previous works analysing NMT outputs, NMT systems are more prone to accuracy errors than fluency errors~\cite{Castilho2017ACQ,Toral17eacl,Klubicka2018,VanBrussel18lrec}.
\item In accordance with automatic evaluation, the total increase of errors between offline and online (last row of \tab{raw}) is higher for \deen~than for \engde~and PA is slightly less impacted by the shift from offline to online. This is possibly due to the fact that TF, a context-based model, is more affected by missing context compared to the position-based convolutional PA.
\item Unlike automatic evaluation where TF slightly outperforms PA, in three out of the four setups in \tab{raw}, PA has less errors than TF.
\item Relative increase of errors between offline and online is larger for fluency than for accuracy, especially for TF.
\item Relative increase of errors is particularly high for addition (ad), word order (wo) and duplication (du)\footnote{
Unwarranted content duplications were marked as duplication errors if fluency was affected or as addition errors otherwise, see annotation guidelines available  at \url{https://github.com/elbayadm/OnlineMT-Evaluation}.
}, the latter being even more problematic for TF. 
\item In line with other error-annotation studies~\cite{Klubicka2018,VanBrussel18lrec,Specia17mt}, mistranslation (mt) is the largest contributor to accuracy errors. Annotating mistranslations is particularly ambiguous leading to a lower inter-annotator agreement.
\item Offline errors with the most consistent gap between PA and TF are duplication and omission in favor of PA and grammar and overly literal in favor of TF. 
\item  More grammar errors are found for \engde~compared to \deen, one reason might be that 
German morphology is richer and more complex than in English, leading to more possibilities for a system to make grammar errors. 
\item Typography errors are more prevalent in \engde~and are highly impacted by latency constraints: most of these typography errors are incorrect punctuation marks (extraneous or missing commas) as well as wrong casing and missing white spaces between compound nouns (producing correct German compounds seems to be especially difficult for online systems). This increase of typography errors suggests that online systems are more literal, as evidenced by the prevalence of incorrect punctuation. 
\end{observations}

\subsection{Fine-grain analysis}
\begin{figure}
\setlength{\tabcolsep}{0pt}
\centering
\begin{subfigure}[b]{\linewidth}
\centering
\small
\begin{tabular}{cc}
    \deen & \engde \\
    \includegraphics[height=3.7cm]{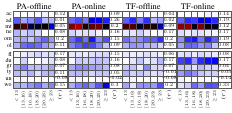} &
    \includegraphics[height=3.8cm]{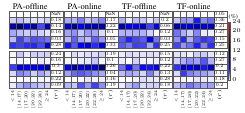} \\
    \vspace{-20pt}
\end{tabular}
\caption{Source length ($\lx$)}\label{fig:srclength}
\end{subfigure}\\
\begin{subfigure}[b]{\linewidth}
\centering
\begin{tabular}{cc}
    \includegraphics[height=3.8cm]{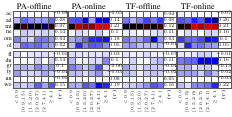} &
    \includegraphics[height=3.8cm]{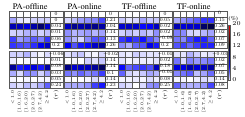} \\
    \vspace{-23pt}
\end{tabular}
\caption{\Ld~($\textrm{LD}(\x,\y)$)}\label{fig:difficulty}
\end{subfigure}
\vspace{-8mm}
\caption{Bucketed segment-level count of errors.}\label{fig:length-diff}
\vspace{-8mm}
\end{figure}

\begin{figure}
\setlength{\tabcolsep}{0pt}
\centering
\begin{subfigure}[b]{\linewidth}
\centering
\small
\begin{tabular}{cc}
    \deen & \engde \\
    \includegraphics[height=4.0cm]{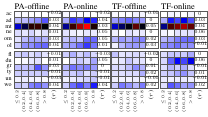} &
    \includegraphics[height=4.0cm]{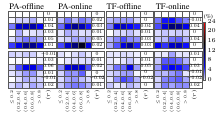} \\
\end{tabular}
\vspace{-10pt}
\caption{Target-side relative position ($t/\lhyp$)}\label{fig:tgt-rpos}
\end{subfigure}
\begin{subfigure}[b]{\linewidth}
\centering
\small
\begin{tabular}{cc}
    \includegraphics[height=4.0cm]{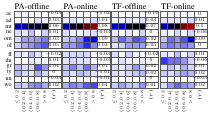} &
    \includegraphics[height=4.0cm]{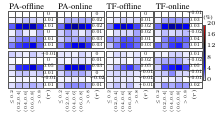} \\
\end{tabular}
\vspace{-10pt}
\caption{Source-side relative position ($j/\lx$)}\label{fig:src-rpos}
\end{subfigure}
\vspace{-8mm}
\caption{Bucketed token-level count of errors.}\label{fig:rpos}
\vspace{-4mm}
\end{figure}

In the following section, we breakdown the annotated set according to the analysis factors (source length ($\lx$), \ld~($\textrm{LD}(\x, \y)$) and relative positions). 
For each of these factors, we bin the annotated segments or tokens and report in Figures~\ref{fig:length-diff} and~\ref{fig:rpos} normalized counts of errors (divided by the total number of tokens in each bin).
An error corresponds to a row of the figure with a heat-map of the normalized counts (read left-to-right with an increasing factor) followed with Pearson's $r$ correlation coefficient.

\mypar{Source length}
In \fig{srclength}, 
\begin{observations}
\item in accordance with the automatic evaluation results of \fig{bleu-buckets}, the relative count of errors per length bucket is positively correlated with  length. 
However, for mistranslation (mt) we observe a peak of the relative count of errors in early bins with a decrease of errors for longer segments. 
This could be attributed to the higher cognitive load in the annotation of longer segments,  
 as it may be easier for annotators to exhaustively label the errors in short segments than in longer ones.\footnote{IAA scores drop in longer segments 
which leads us to speculate that it is difficult to annotate longer segments.}
The ease of annotating short segments can also be attributed to their fluency; since these segments have fewer fluency errors, labeling mistranslated words is less ambiguous.
\item Addition (ad) errors in \deen~online systems are considerably more correlated with the source length compared to the offline systems (PA: 0.01$\sto$0.26 and TF: 0.02$\sto$0.19). This shows that the increase in these errors is mostly located in longer segments.
\item Omission (om) and duplication (du) errors, more problematic for TF, have a higher correlation with the source length in \engde~(du: 0.12$\sto$0.27 and om: 0.03$\sto$0.15) but not in \deen. 
\end{observations}
\mypar{Difficulty}
\fig{difficulty} shows that 
\begin{observations}
\item even in offline systems, the relative error count is positively correlated with \ld. This is particularly noticeable for word order (wo) errors.
\item For online systems, additions (ad) and omissions (om) are particularly correlated with the \ld. These errors are the system's solution to deal with missing context.
\item Although duplication (du) can also be thought of as a solution to missing context, it is more correlated with the source length than with the difficulty.
\end{observations}

\mypar{Relative position}
Unlike length and \ld, analysis of relative position is based on token-wise labels.
In \fig{rpos}, we observe that 
\begin{observations}
\item in online \deen~systems, most omission (om) and mistranslation (mt) errors concern final source positions and occur near the end of the translation. 
The high prevalence of these error types in final source positions confirms the well-known difficulty of German sentence-final structures. 
Note that sentence-final errors on the source side (De) do not necessarily lead to sentence-final errors in the target (En), since the structural differences between the two languages require reordering operations.
\item Duplication (du) errors in TF, on the other hand, affect initial source positions near the end of the translation. This means that TF circles back to the beginning of the source to pad the length of the hypothesis.
\item Compared to \deen, the errors in \engde~systems are well spread across the positions. This is likely due to the fact that decoding English (verb-medial) from German (verb-final) is more exposed to the issue of missing context from final positions.
\item Having more mistranslation (mt) errors near the end of the hypotheses in \deen~is probably due to the ambiguity of mistranslation errors (the most frequent type). An omission followed by an addition is logically interpreted as a mistranslation (see annotation (d) of \tab{examples} where \emph{taking} could also be labeled a mistranslation).
\end{observations}

\begin{table}
\small
\centering
\setlength{\tabcolsep}{2.5pt}
% Picking examples from de-en annotated by A2 (Neumanns)
\begin{tabular}{cll}
\toprule
%Source & Es geht darum, die Nuancen der Sprache zu verstehen.\\
%Reference &  It's about the nuance of human language.\\
\multirow{8}{*}{\rotatebox{90}{Example 1}}
           & \multirow{2}{*}{(a)} 
           & Source: Es geht darum, die Nuancen der Sprache zu verstehen. \\
           && PA-offline:  It's about understanding the nuances of language. \\
           \cmidrule{2-3}
           &\multirow{2}{*}{(b)}
           & Source: Es geht darum, die Nuancen der Sprache\ulc{red}{zu}{ol}\ulc{p1purple}{ verstehen}{mt, wo}.\\
           && PA-online: It's about the nuances of language\ulc{red}{to}{ol}\ulc{p1purple}{understand}{mt, wo}.\\
            \cmidrule{2-3}
           &\multirow{2}{*}{(c)}
           & Source: Es geht darum, die Nuancen der Sprache zu verstehen. \\
           && TF-offline: It's about understanding the nuances of language. \\
           \cmidrule{2-3}
           &\multirow{2}{*}{(d)} 
           & Source: Es geht darum, die Nuancen der Sprache zu \ulc{blue}{verstehen}{wo}. \\
           && Tf-online: It's about\ulc{red}{taking}{ad} the nuances of language,\ulc{blue}{understanding}{wo}\ulc{blue}{the}{du}\ulc{blue}{nuances}{du}.\\
\midrule
%Source & Und wenn wir dies für Rohdaten machen können warum nicht auch für Inhalte selbst?\\
%Reference & And if we can do this for raw data, why not do it for content as well?\\
\multirow{8}{*}{\rotatebox{90}{Example 2}}
           &\multirow{2}{*}{(e)} 
           & Source: Und wenn wir dies für Rohdaten machen können warum nicht auch für Inhalte\ulc{red}{selbst}{ol}?\\
           && PA-offline: And if we can do that for raw data, why not for content\ulc{red}{itself}{ol}?\\
           \cmidrule{2-3}
           &\multirow{2}{*}{(f)}
           & Source: Und wenn wir dies für Rohdaten machen\ulc{red}{können}{om} warum nicht auch für Inhalte\ulc{red}{selbst?}{ol}\\
           && PA-online: And if we\ulc{red}{}{om} do this for raw data, why not content\ulc{red}{itself?}{ol}\\
           \cmidrule{2-3}
           &\multirow{2}{*}{(g)} 
           & Source: Und wenn wir dies für Rohdaten machen können warum nicht auch für Inhalte\ulc{red}{selbst?}{ol}\\
           && TF-offline: And if we can do this for raw data, why not for content\ulc{red}{itself}{ol}?\\
           \cmidrule{2-3}
           &\multirow{2}{*}{(h)} 
           & Source: Und wenn wir dies für Rohdaten machen können warum nicht\ulc{red}{auch}{om} für Inhalte\ulc{red}{selbst?}{ol}\\
           && TF-online : And if we could do this for raw data, why not do it\ulc{red}{}{om} for content \ulc{red}{itself?}{ol}\\
\bottomrule
\end{tabular}
%% ------ Example 1:
%% ID=6234. AnnotID=151
%% ------ Example 2:
%% ID=2294 AnnotID=9

\vspace{-2mm}
\caption{Example annotations from \deen. Accuracy errors are in red and fluency in blue.
}\label{tab:examples}
\vspace{-6mm}
\end{table}

\subsection{Agreement of the automatic metrics with human annotation}
In \tab{correlation} we evaluate the correlation (Pearson's $r$) between the automatic metrics and the human judgement by proxy of the error count. Unlike in \sect{automatic} where we evaluate corpus-level BLEU, in this section we use sentence-level BLEU denoted with SentBLEU.
We highlight the following observations:
\begin{observations}
\item On average, the model's confidence ($\conf$) and BERTScore have the highest correlation with the annotated errors in \deen. In \engde~BERTScore is less correlated with the human evaluation, this is likely due to the use of a smaller German model (\emph{base} instead of the English \emph{large}). The correlation is higher for accuracy errors.
\item  With its high correlation with errors, the confidence might be used to efficiently decide when to read and when to write instead of following a deterministic decoding path such as \waitk.  \cite{Cho16arxiv,Liu20arxiv}
\item TER (the normalized number of edits required to get from the hypothesis to the reference) is a better indicator of quality than BLEU and METEOR in online systems. It is particularly correlated with addition (ad) and duplication (du) errors frequent online.
\item Although omissions (om) in \engde~are negatively correlated with confidence and can thus be avoided with a well tuned decoding algorithm, this is not the case for \deen~(verb-final to verb-medial). This means that the model omits tokens with a high confidence and is unable to predict that context is missing from the source.
\end{observations}

\begin{table}
    \scriptsize
    \centering
    \setlength{\tabcolsep}{2.5pt}
    \settowidth{\rotheadsize}{SentBLEU}
\begin{tabular}{r|HrrrrrHHHH|HrrrrrHHHH|HrrrrrHHHH|HrrrrrHHHH}
\toprule
\multicolumn{41}{c}{\bf\deen}\\
\toprule
& \multicolumn{10}{c|}{PA-offline} & \multicolumn{10}{c|}{PA-online} & \multicolumn{10}{c|}{TF-offline} & \multicolumn{10}{c}{TF-online} \\
\hline
%& LD & \rothead{\scriptsize TER} & \rothead{\scriptsize SentBLEU} & \rothead{\scriptsize METEOR} & \rothead{\scriptsize BERTScore} & \rothead{\scriptsize$\conf$} & lx & ly & lhyp & \rothead{\scriptsize$\ratio$}
%& LD & \rothead{\scriptsize TER} & \rothead{\scriptsize SentBLEU} & \rothead{\scriptsize METEOR} & \rothead{\scriptsize BERTScore} & \rothead{\scriptsize$\conf$} & lx & ly & lhyp & \rothead{\scriptsize$\ratio$}
%& LD & \rothead{\scriptsize TER} & \rothead{\scriptsize SentBLEU} & \rothead{\scriptsize METEOR} & \rothead{\scriptsize BERTScore} & \rothead{\scriptsize$\conf$} & lx & ly & lhyp & \rothead{\scriptsize$\ratio$}
%& LD & \rothead{\scriptsize TER} & \rothead{\scriptsize SentBLEU} & \rothead{\scriptsize METEOR} & \rothead{\scriptsize BERTScore} & \rothead{\scriptsize$\conf$} & lx & ly & lhyp & \rothead{\scriptsize$\ratio$}\\
& LD & T & sB & M & B & p & lx & ly & lhyp & ratio 
& LD & T & sB & M & B & p & lx & ly & lhyp & ratio 
& LD & T & sB & M & B & p & lx & ly & lhyp & ratio 
& LD & T & sB & M & B & p & lx & ly & lhyp & ratio \\
\toprule
%ac & -0.09 & -0.02 & -0.05 & -0.05 & -0.03 & -0.07 & 0.12 & 0.15 & 0.10 & -0.04 & -0.05 & 0.00 & -0.06 & -0.10 & -0.01 & -0.04 & 0.09 & 0.13 & 0.07 & -0.04 & 0.04 & -0.04 & 0.02 & 0.01 & -0.02 & -0.13 & 0.04 & 0.07 & 0.07 & 0.06 & -0.07 & 0.04 & -0.07 & -0.08 & -0.09 & -0.11 & 0.14 & 0.14 & 0.11 & -0.05 \\ 
ad & 0.08 & 0.26 & -0.16 & -0.19 & -0.28 & \bf-0.42 & 0.01 & 0.01 & 0.21 & 0.39 
   & 0.14 & \bf0.38 & -0.28 & -0.27 & \bf-0.35 & \bf-0.40 & 0.26 & 0.15 & 0.38 & 0.24 
   & 0.08 & \bf0.43 & -0.28 & -0.28 & \bf-0.33 & \bf-0.46 & 0.02 & 0.00 & 0.20 & 0.35 
   & 0.26 & \bf0.39 & \bf-0.33 & -0.26 &  \bf-0.38 & \bf-0.41 & 0.19 & 0.10 & 0.32 & 0.23 \\ 
mt & 0.07 & \bf0.36 & \bf-0.41 & \bf-0.36 & \bf-0.54 & \bf-0.45 & 0.05 & 0.05 & 0.13 & 0.17 
   & 0.21 & \bf0.41 & \bf-0.42 & \bf-0.42 & \bf-0.51 & \bf-0.46 & 0.20 & 0.10 & 0.30 & 0.22 
   & 0.12 & \bf0.42 & \bf-0.41 & \bf-0.44 & \bf-0.60 & \bf-0.54 & 0.09 & 0.05 & 0.20 & 0.17 
   & 0.21 & \bf0.40 & \bf-0.41 & \bf-0.40 & \bf-0.54 & \bf-0.50 & 0.12 & 0.10 & 0.24 & 0.23 \\ 
ne & 0.14 & 0.09 & -0.06 & -0.07 & -0.04 & -0.17 & 0.08 & 0.07 & 0.14 & 0.11 
   & 0.10 & 0.13 & -0.04 & -0.10 & -0.09 & -0.16 & 0.16 & 0.07 & 0.21 & 0.05 
   & 0.11 & 0.10 & -0.08 & -0.07 & -0.10 & -0.04 & 0.17 & 0.22 & 0.23 & 0.08 
   & 0.16 & \bf0.30 & -0.23 & -0.19 & -0.21 & -0.21 & 0.17 & 0.01 & 0.30 & 0.22 \\ 
om & 0.04 & 0.10 & -0.11 & -0.14 & -0.21 & -0.20 & 0.20 & 0.12 & 0.09 & -0.19 
   & 0.18 & 0.06 & -0.11 & -0.06 & -0.24 & -0.10 & 0.15 & 0.14 & 0.01 & -0.31 
   & 0.03 & 0.07 & -0.13 & -0.08 & -0.10 & -0.14 & 0.20 & 0.14 & 0.10 & -0.19 
   & 0.10 & 0.06 & -0.11 & -0.07 & -0.21 & -0.10 & 0.19 & 0.14 & 0.11 & -0.15 \\ 
ol & 0.02 & 0.12 & -0.12 & -0.16 & -0.07 & -0.07 & 0.11 & 0.10 & 0.09 & -0.06 
   & 0.05 & 0.12 & -0.11 & -0.12 & -0.15 & -0.10 & 0.09 & 0.07 & 0.10 & 0.03 
   & -0.01 & 0.14 & -0.12 & -0.17 & -0.15 & -0.15 & 0.05 & 0.07 & 0.10 & 0.09 
   & 0.05 & 0.26 & -0.20 & -0.22 & -0.23 & -0.20 & 0.08 & 0.01 & 0.17 & 0.15 \\ 
\hline
ac+ & 0.10 & \bf0.43 & \bf-0.43 & \bf-0.44 & \bf-0.62 & \bf-0.58 & 0.10 & 0.09 & 0.20 & 0.21 
    & 0.28 & \bf0.49 & \bf-0.48 & \bf-0.46 & \bf-0.63 & \bf-0.56 & 0.31 & 0.19 & 0.39 & 0.16 
    & 0.14 & \bf0.53 & \bf-0.48 & \bf-0.49 & \bf-0.63 & \bf-0.66 & 0.14 & 0.09 & 0.27 & 0.21 
    & 0.29 & \bf0.50 & \bf-0.48 & \bf-0.43 & \bf-0.62 & \bf-0.58 & 0.24 & 0.15 & 0.36 & 0.24 \\ 
\hline
fl & -0.01 & -0.01 & -0.01 & 0.05 & 0.02 & 0.06 & 0.17 & 0.17 & 0.19 & -0.02 
   & 0.03 & 0.03 & -0.04 & 0.02 & -0.05 & -0.00 & 0.15 & 0.07 & 0.12 & -0.07 
   & -0.07 & 0.05 & -0.10 & -0.05 & -0.08 & -0.01 & 0.06 & 0.06 & 0.03 & \bf-0.09 
   & -0.01 & -0.00 & -0.04 & -0.01 & 0.02 & 0.00 & 0.08 & 0.09 & 0.07 & -0.04 \\ 
du & 0.14 & 0.09 & -0.06 & -0.07 & -0.04 & -0.17 & 0.08 & 0.07 & 0.14 & 0.11 
   & 0.10 & 0.13 & -0.04 & -0.10 & -0.09 & -0.16 & 0.16 & 0.07 & 0.21 & 0.05 
   & 0.11 & 0.10 & -0.08 & -0.07 & -0.10 & -0.04 & 0.17 & 0.22 & 0.23 & 0.08 
   & 0.16 & \bf0.30 & -0.23 & -0.19 & -0.21 & -0.21 & 0.17 & 0.01 & 0.30 & 0.22 \\ 
gr & 0.10 & 0.08 & -0.12 & -0.08 & -0.08 & -0.01 & 0.07 & 0.07 & 0.10 & 0.05 
   & -0.06 & 0.02 & -0.11 & 0.04 & -0.09 & -0.10 & 0.08 & 0.09 & 0.08 & 0.02 
   & -0.04 & 0.04 & -0.11 & -0.02 & -0.09 & -0.05 & 0.07 & 0.10 & 0.10 & 0.05 
   & -0.13 & -0.02 & -0.04 & 0.04 & -0.03 & -0.10 & 0.04 & 0.03 & 0.04 & 0.02 \\ 
ty & -0.04 & -0.01 & -0.06 & 0.04 & 0.02 & 0.02 & 0.11 & 0.08 & 0.09 & -0.01 
   & -0.03 & 0.05 & -0.06 & -0.04 & -0.05 & 0.09 & 0.05 & 0.05 & 0.07 & 0.08 
   & -0.02 & -0.04 & 0.02 & 0.01 & -0.01 & 0.11 & -0.01 & 0.03 & -0.00 & 0.03 
   & -0.02 & 0.07 & -0.08 & -0.15 & -0.04 & 0.04 & -0.05 & -0.02 & -0.02 & 0.08 \\ 
%un & -0.01 & 0.11 & -0.05 & -0.09 & -0.18 & -0.13 & -0.09 & -0.07 & -0.06 & 0.05 
    %& 0.08 & 0.21 & -0.12 & -0.20 & -0.15 & -0.22 & -0.02 & -0.04 & -0.04 & -0.03 
    %& 0.05 & 0.20 & -0.10 & -0.11 & -0.09 & -0.14 & -0.08 %& -0.12 & -0.06 & 0.03 
    %& -0.03 & 0.29 & -0.12 & -0.13 & -0.28 & -0.26 & -0.14 & -0.11 & -0.05 & 0.30 \\ 
wo & 0.15 & 0.07 & -0.05 & -0.06 & -0.09 & -0.10 & 0.15 & 0.20 & 0.19 & 0.06 
   & 0.19 & 0.13 & -0.10 & 0.07 & -0.17 & 0.01 & 0.30 & 0.20 & 0.28 & -0.05 
   & 0.16 & 0.17 & -0.19 & -0.17 & -0.17 & -0.03 & 0.20 & 0.11 & 0.15 & -0.08 
   & 0.22 & 0.11 & -0.09 & -0.10 & -0.13 & -0.02 & 0.33 & 0.22 & 0.30 & -0.08 \\ 
\hline
fl+ & 0.17 & 0.11 & -0.14 & -0.11 & -0.11 & -0.11 & 0.23 & 0.25 & 0.29 & 0.09 
    & 0.15 & 0.20 & -0.17 & -0.14 & -0.23 & -0.09 & 0.36 & 0.24 & 0.36 & 0.01 
    & 0.11 & 0.17 & -0.21 & -0.16 & -0.20 & -0.02 & 0.22 & 0.19 & 0.22 & -0.01 
    & 0.18 & \bf0.30 & -0.24 & -0.24 & -0.24 & -0.18 & 0.27 & 0.11 & 0.36 & 0.16 \\ 
\hline
ac+fl & 0.15 & \bf0.41 & \bf-0.43 & \bf-0.42 & \bf-0.58 & \bf-0.54 & 0.18 & 0.18 & 0.29 & 0.21 
      & 0.30 & \bf0.50 & \bf-0.48 & \bf-0.45 & \bf-0.64 & \bf-0.51 & 0.41 & 0.26 & 0.48 & 0.14 
      & 0.17 & \bf0.54 & \bf-0.51 & \bf-0.50 & \bf-0.64 & \bf-0.59 & 0.21 & 0.16 & 0.32 & 0.18 
      & 0.31 & \bf0.54 & \bf-0.50 & \bf-0.46 & \bf-0.61 & \bf-0.55 & 0.32 & 0.17 & 0.46 & 0.26 \\ 
\bottomrule
\multicolumn{41}{c}{\bf\engde}\\
\toprule
ad & -0.00 & 0.06 & -0.11 & -0.10 & -0.14 & \bf-0.23 & 0.03 & 0.10 & 0.18 & 0.26 
   & 0.21 & 0.22 & -0.24 & -0.25 & -0.23 & \bf-0.30 & 0.14 & 0.04 & 0.17 & 0.11 
   & 0.05 & 0.18 & -0.09 & -0.14 & -0.19 & \bf-0.34 & 0.03 & 0.03 & 0.20 & 0.33 
   & 0.15 & 0.22 & -0.24 & -0.21 & -0.25 & \bf-0.48 & 0.29 & 0.11 & 0.36 & 0.22 \\ 

mt & 0.04 & 0.14 & -0.27 & -0.21 & -0.28 & \bf-0.43 & 0.13 & 0.16 & 0.14 & 0.05 
   & 0.04 & 0.05 & -0.24 & -0.15 & -0.20 & \bf-0.56 & 0.26 & 0.25 & 0.22 & -0.08 
   & 0.02 & 0.08 & -0.22 & -0.17 & -0.18 & \bf-0.39 & 0.12 & 0.07 & 0.06 & -0.11 
   & 0.26 & 0.22 & -0.26 & -0.26 & -0.31 & \bf-0.44 & 0.24 & 0.16 & 0.21 & -0.00 \\ 

ne & 0.01 & -0.02 & 0.03 & 0.03 & -0.01 & 0.02 & 0.12 & 0.15 & 0.16 & 0.07 
   & 0.14 & 0.17 & -0.13 & 0.09 & -0.06 & -0.14 & 0.12 & -0.03 & 0.10 & -0.00 
   & 0.02 & 0.26 & -0.21 & -0.25 & -0.19 & -0.27 & 0.05 & -0.01 & 0.12 & 0.11 
   & 0.02 & 0.25 & -0.18 & -0.12 & -0.11 & -0.28 & 0.17 & 0.02 & 0.27 & 0.25 \\ 

om & 0.06 & 0.07 & -0.18 & -0.15 & -0.20 & \bf-0.34 & 0.19 & 0.18 & 0.03 & -0.27 
   & 0.13 & 0.17 & -0.25 & -0.24 & -0.25 & \bf-0.35 & 0.13 & 0.07 & 0.05 & -0.17 
   & -0.09 & 0.11 & -0.21 & -0.15 & -0.16 & -0.22 & 0.15 & 0.14 & 0.03 & -0.19 
   & 0.07 & 0.10 & -0.25 & -0.17 & -0.24 & \bf-0.40 & 0.19 & 0.21 & 0.15 & -0.05 \\ 

ol & 0.20 & 0.17 & \bf-0.31 & -0.22 & \bf-0.27 & -0.25 & 0.31 & 0.27 & 0.28 & -0.00 
   & 0.26 & 0.18 & -0.28 & -0.24 & \bf-0.34 & -0.27 & 0.33 & 0.22 & 0.33 & 0.04 
   & 0.20 & 0.12 & -0.18 & -0.11 & -0.15 & -0.10 & 0.20 & 0.14 & 0.25 & 0.13 
   & 0.09 & 0.11 & -0.16 & -0.16 & -0.26 & -0.24 & 0.26 & 0.18 & 0.25 & 0.02 \\ 
\hline
ac+ & 0.13 & 0.20 & \bf-0.39 & -0.29 & \bf-0.41 & \bf-0.58 & 0.28 & 0.30 & 0.24 & -0.02 
    & 0.23 & 0.21 & \bf-0.40 & \bf-0.32 & \bf-0.41 & \bf-0.66 & 0.38 & 0.27 & 0.33 & -0.06 
    & 0.10 & 0.20 & \bf-0.35 & -0.26 & \bf-0.32 & \bf-0.51 & 0.23 & 0.18 & 0.23 & 0.02 
    & 0.22 & 0.25 & \bf-0.36 & \bf-0.33 & \bf-0.43 & \bf-0.66 & 0.34 & 0.23 & 0.34 & 0.07 \\ 
\hline
fl & -0.06 & 0.00 & -0.01 & -0.02 & -0.03 & -0.13 & 0.16 & 0.18 & 0.24 & 0.17 
   & -0.02 & 0.02 & -0.08 & -0.06 & -0.12 & -0.12 & 0.17 & 0.16 & 0.19 & 0.08 
   & -0.01 & 0.01 & -0.04 & -0.00 & -0.03 & -0.10 & 0.08 & 0.09 & 0.15 & 0.16 
   & -0.03 & -0.07 & 0.06 & 0.09 & 0.08 & 0.00 & 0.12 & 0.10 & 0.12 & 0.03 \\ 

du & 0.01 & -0.02 & 0.03 & 0.03 & -0.01 & 0.02 & 0.12 & 0.15 & 0.16 & 0.07 
   & 0.14 & 0.17 & -0.13 & 0.09 & -0.06 & -0.14 & 0.12 & -0.03 & 0.10 & -0.00 
   & 0.02 & 0.26 & -0.21 & -0.25 & -0.19 & -0.27 & 0.05 & -0.01 & 0.12 & 0.11 
   & 0.02 & 0.25 & -0.18 & -0.12 & -0.11 & -0.28 & 0.17 & 0.02 & 0.27 & 0.25 \\ 

gr & 0.08 & -0.00 & -0.13 & -0.03 & -0.04 & -0.12 & 0.30 & 0.30 & 0.30 & 0.04 
   & 0.14 & 0.03 & -0.19 & -0.11 & -0.10 & -0.28 & 0.30 & 0.25 & 0.26 & -0.07 
   & 0.18 & 0.01 & -0.08 & -0.03 & -0.05 & -0.01 & 0.24 & 0.24 & 0.22 & -0.00 
   & 0.07 & 0.11 & -0.19 & -0.10 & -0.08 & -0.17 & 0.23 & 0.17 & 0.20 & -0.01 \\ 

ty & 0.03 & 0.10 & -0.06 & -0.04 & -0.09 & -0.07 & 0.08 & 0.13 & 0.12 & 0.07 
   & 0.10 & -0.01 & 0.04 & 0.07 & 0.03 & 0.00 & 0.12 & 0.02 & 0.04 & -0.17 
   & -0.01 & 0.09 & -0.08 & -0.06 & -0.05 & -0.07 & 0.07 & 0.09 & 0.13 & 0.10 
   & 0.05 & -0.00 & -0.02 & 0.03 & 0.02 & -0.06 & 0.11 & 0.06 & 0.11 & 0.01 \\ 

wo & 0.24 & 0.23 & -0.24 & -0.21 & -0.26 & -0.13 & 0.14 & 0.02 & 0.09 & -0.07 
   & 0.23 & 0.04 & -0.14 & -0.08 & -0.17 & -0.18 & 0.25 & 0.20 & 0.19 & -0.11 
   & 0.25 & 0.25 & -0.18 & -0.18 & -0.26 & -0.09 & 0.11 & 0.03 & 0.18 & 0.12 
   & 0.08 & 0.19 & -0.21 & -0.18 & -0.20 & -0.20 & 0.16 & 0.10 & 0.20 & 0.13 \\ 

\hline

fl+ & 0.14 & 0.11 & -0.20 & -0.11 & -0.16 & -0.19 & 0.35 & 0.33 & 0.37 & 0.09 
    & 0.25 & 0.08 & -0.22 & -0.13 & -0.18 & \bf-0.32 & 0.40 & 0.29 & 0.33 & -0.13 
    & 0.20 & 0.17 & -0.19 & -0.15 & -0.19 & -0.16 & 0.27 & 0.23 & 0.32 & 0.14 
    & 0.10 & 0.23 & -0.27 & -0.16 & -0.17 & \bf-0.31 & 0.32 & 0.19 & 0.35 & 0.14 \\ 
\hline

ac+fl & 0.17 & 0.21 & \bf-0.39 & -0.28 & \bf-0.39 & \bf-0.53 & 0.40 & 0.41 & 0.39 & 0.04 
      & \bf0.30 & 0.18 & \bf-0.39 & -0.29 & \bf-0.38 & \bf-0.63 & 0.49 & 0.35 & 0.41 & -0.12 
      & 0.19 & 0.24 & \bf-0.35 & -0.27 & \bf-0.33 & \bf-0.44 & 0.31 & 0.26 & 0.35 & 0.09 
      & 0.21 & \bf0.30 & \bf-0.39 & \bf-0.31 & \bf-0.38 & \bf-0.62 & 0.42 & 0.27 & 0.44 & 0.13 \\ 
\bottomrule
\end{tabular}

\vspace{-1mm}
\caption{Correlation of the automatic metrics 
        (T: TER, sB: SentBLEU, M: METEOR, B: BERTScore, p: the model's confidence $\conf$) 
        with the error counts (Pearson's $r$). Correlations with $|r|\geq 0.3$ in bold.
    }\label{tab:correlation}
\vspace{-3mm}
\end{table}

\section{Conclusion}\label{sec:conclusion} 
We have conducted an evaluation of offline and online NMT systems for spoken language translation. 
Our aim was to shed light on the strengths and weaknesses of \waitk~decoding under two different architectures, Transformer and Pervasive Attention.
We found that Transformer models are strongly affected by the shift to online decoding with a significant increase in fluency errors, most of which are duplications. PA on the other hand accrues less degradation in online decoding.
Our error analysis shows that
translation quality in online models can be potentially improved by making read/write decisions based on the model's confidence in order to filter out avoidable additions, mistranslations and duplications.
The syntactic asymmetry between German and English remains a challenge for deterministic online decoding.
A more detailed analysis of syntactically required long-distance reorderings is left for future work. In this regard, indicators such as lagging difficulty or relative position are more informative for online translation.
Another line of future work is to assess to what extent these findings carry over to ASR outputs

\mypar{Acknowledgements}
We would like to thank the annotators for their valuable contributions.
\ifcolingfinal
This work partially results from a research collaboration co-financed by the University of Innsbruck's France Focus.
\else
\fi

\FloatBarrier
\bibliographystyle{coling}
\bibliography{references}

\newpage
\appendix
\section{Inter-annotator agreement}\label{appendix:agreement}
\begin{figure}[H]
    \centering
    \begin{subfigure}[b]{.24\linewidth}
    \includegraphics[height=4cm]{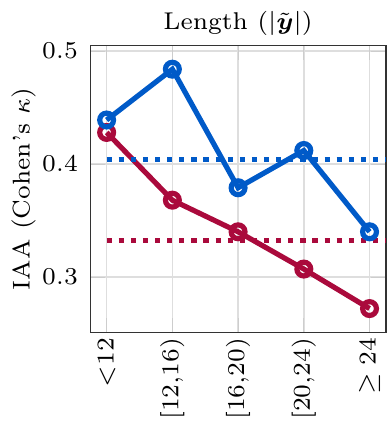}
    \caption{Length  - $\kappa$}\label{fig:agree-length}
    \end{subfigure}%
    \begin{subfigure}[b]{.35\linewidth}
    \includegraphics[height=4cm]{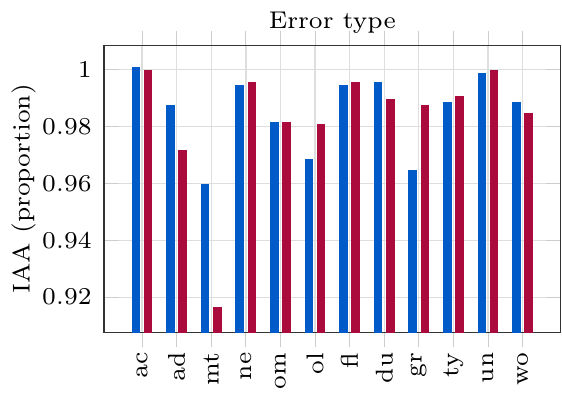}
    \caption{Error type - agreement proportion}\label{fig:agree-absolute}
    \end{subfigure}
    \begin{subfigure}[b]{.35\linewidth}
    \includegraphics[height=4cm]{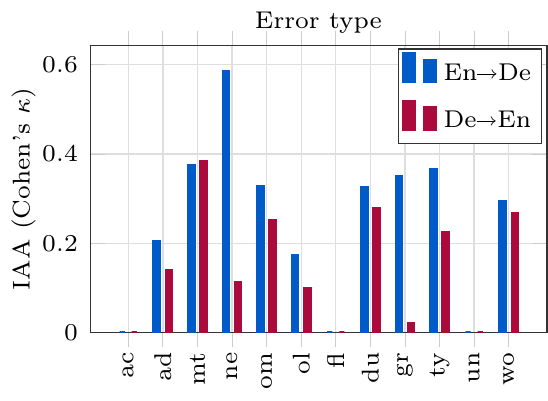}
    \caption{Error type - $\kappa$}\label{fig:agree-kappa}
    \end{subfigure}%
    \caption{IAA measured with Cohen's kappa or as agreement proportion without chance correction. The left panel shows the IAA per hypothesis length and the two right panels breakdown the agreement per error type.}\label{fig:agree}
\end{figure}

\mypar{Length}
To study whether long segments are more challenging to annotate, we measure inter-annotator agreement (IAA) in buckets of source length.
Similar to prior studies~\cite{Flammia95sct,Stymne12lrec,Bojar11wmt}, we found that the length of the sequence has a negative correlation with the agreement possibly because of the increasing cognitive load (see \fig{agree-length}).

\mypar{Error type} To assess the ambiguity of the error types in our study, we measure binary agreements for each error in the typology. In this setup, we consider the task of annotating each error as a binary classification. Without chance correction (\fig{agree-absolute}), mistranslation (mt) error is the one with the highest disagreement possibly because of its ambiguity. Agreement on rare errors such as accuracy (ac) and unintelligible (un) is zeroed out after chance correction.

\end{document}